\documentclass[letterpaper, 10 pt, conference]{style/ieeeconf}  

\makeatletter
\let\NAT@parse\undefined
\makeatother

\usepackage{dblfloatfix}

\usepackage[numbers,sectionbib,sort&compress]{natbib}
\usepackage{bm}
\usepackage{gensymb}
\usepackage{graphicx}
\usepackage{amsmath}
\usepackage{amssymb}
\usepackage{algorithm}
\usepackage[noend]{algpseudocode}
\usepackage{subcaption}
\usepackage{amsfonts}
\usepackage{siunitx}
\usepackage{booktabs}
\usepackage{makecell}
\usepackage{multirow}
\usepackage{upgreek}
\usepackage[font=small]{caption}
\usepackage[export]{adjustbox}
\usepackage{tikz}
\usepackage{sidecap} \sidecaptionvpos{figure}{c}
\DeclareMathOperator*{\argmax}{argmax}

\usepackage{hyperref}
\hypersetup{colorlinks,breaklinks,
linkcolor=[rgb]{0.5,0.,0.},
citecolor=[rgb]{0.000,0.427,0.173},
urlcolor=[rgb]{0.031,0.318,0.612}}

\usepackage[nameinlink]{cleveref}
\usepackage{color}
\definecolor{CommentPink}{rgb}{1,0.2,0.5}
\definecolor{CommentBlue}{rgb}{0,0,1}
\definecolor{CommentGreen}{rgb}{0,1,0}

\Crefname{section}{Sec.}{Sec.}
\Crefname{figure}{Fig.}{Fig.}
\Crefname{equation}{Eq.}{Eq.}

\usepackage[printonlyused,withpage,nolist,nohyperlinks]{acronym}
\begin{acronym}

\acro{2D}{two-dimensional}

\acro{3D}{three-dimensional}

\acro{AHRS}{attitude and heading reference system}

\acro{AUV}{autonomous underwater vehicle}

\acro{COMA}{counterfactual multi-agent policy gradients}

\acro{CPP}{Chinese Postman Problem}

\acro{DoF}{degree of freedom}
\acrodefplural{DoF}[DoFs]{degrees of freedom}

\acro{DVL}{Doppler velocity log}

\acro{FSM}{finite state machine}

\acro{IMU}{inertial measurement unit}

\acro{LBL}{Long Baseline}

\acro{MCM}{mine countermeasures}

\acro{MDP}{Markov decision process}
\acrodefplural{MDP}[MDPs]{Markov decision processes}

\acro{MCTS}{Monte Carlo Tree Search}

\acro{POMDP}{Partially Observable Markov Decision Process}
\acrodefplural{POMDP}[POMDPs]{Partially Observable Markov Decision Processes}

\acro{PRM}{Probabilistic Roadmap}
\acrodefplural{PRM}[PRM]{Probabilistic Roadmaps}

\acro{RL}{Reinforcement learning}

\acro{ROI}{region of interest}
\acrodefplural{ROI}[ROIs]{regions of interest}

\acro{ROS}{Robot Operating System}

\acro{ROV}{remotely operated vehicle}

\acro{RRT}{Rapidly-exploring Random Tree}
\acrodefplural{RRT}[RRTs]{Rapidly-exploring Random Trees}

\acro{SLAM}{Simultaneous Localisation and Mapping}

\acro{SSE}{sum of squared errors}

\acro{STOMP}{Stochastic Trajectory Optimization for Motion Planning}

\acro{TRN}{Terrain-Relative Navigation}

\acro{UAV}{unmanned aerial vehicle}
\acrodefplural{UAV}[UAVs]{unmanned aerial vehicles}

\acro{USBL}{Ultra-Short Baseline}

\acro{IPP}{informative path planning}

\acro{FoV}{field of view}
\acrodefplural{FoV}[FoVs]{fields of view}

\acro{CDF}{cumulative distribution function}

\acro{ML}{maximum likelihood}

\acro{RMSE}{Root Mean Squared Error}
\acro{MLL}{Mean Log Loss}

\acro{GP}{Gaussian Process}
\acrodefplural{GP}[GPs]{Gaussian Processes}

\acro{KF}{Kalman Filter}

\acro{IP}{Interior Point}
\acro{BO}{Bayesian Optimization}

\acro{SE}{squared exponential}

\acro{UI}{uncertain input}

\acro{MCL}{Monte Carlo Localisation}
\acro{AMCL}{Adaptive Monte Carlo Localisation}

\acro{SSIM}{Structural Similarity Index}

\acro{MAE}{Mean Absolute Error}
\acro{RMSE}{Root Mean Squared Error}
\acro{AUSE}{Area Under the Sparsification Error curve}

\end{acronym}

\IEEEoverridecommandlockouts                           

\title{Multi-UAV Adaptive Path Planning Using Deep Reinforcement Learning}

\author{Jonas Westheider, Julius Rückin, Marija Popovi\'{c}
\thanks{This work was funded by the Deutsche Forschungsgemeinschaft (DFG, German Research Foundation) under Germany's Excellence Strategy - EXC 2070 – 390732324.
Authors are with the Cluster of Excellence PhenoRob, Institute of Geodesy and Geoinformation, University of Bonn.
Corresponding: \texttt{jwesthei@uni-bonn.de}.}%
}

\begin{document}

\maketitle

\begin{abstract}

Efficient aerial data collection is important in many remote sensing applications.
In large-scale monitoring scenarios, deploying a team of unmanned aerial vehicles (UAVs) offers improved spatial coverage and robustness against individual failures.
However, a key challenge is cooperative path planning for the UAVs to efficiently achieve a joint mission goal.
We propose a novel multi-agent informative path planning approach based on deep reinforcement learning for adaptive terrain monitoring scenarios using UAV teams.
We introduce new network feature representations to effectively learn path planning in a 3D workspace.
By leveraging a counterfactual baseline, our approach explicitly addresses credit assignment to learn cooperative behaviour. Our experimental evaluation shows improved planning performance, i.e. maps regions of interest more quickly, with respect to non-counterfactual variants.
Results on synthetic and real-world data show that our approach has superior performance compared to state-of-the-art non-learning-based methods, while being transferable to varying team sizes and communication constraints.

\end{abstract}

\section{Introduction} \label{S:intro}

\acused{UAV}
Efficient aerial data collection is crucial for mapping and monitoring phenomena on the Earth's surface.
Unmanned aerial vehicles (UAVs) provide a flexible, labour-, and cost-efficient solution for remote sensing applications such as precision agriculture~\citep{Albani2018,Carbone2021,Maes2019}, wildlife conservation~\citep{Bondi2018}, and search and rescue~\citep{Goodrich2008,Hayat2017}.
For monitoring large terrains, replacing a single \ac{UAV} with a multi-\ac{UAV} system can improve spatial coverage, versatility, and robustness to individual failures at lower overall cost~\citep{Yan2013}.
However, to fully unlock its potential, a key challenge is planning \ac{UAV} paths cooperatively in complex environments, given on-board constraints on runtime efficiency and communication.

This paper addresses active data collection using a team of \acp{UAV} in terrain monitoring scenarios. Our goal is to map an initially unknown, non-homogeneous binary target variable of interest on a 2D terrain, e.g. crop infestations in an agricultural scenario or to-be-rescued victims in a disaster scenario, using image measurements taken by the \acp{UAV}.
We tackle the problem of multi-agent \textit{\ac{IPP}}: we plan information-rich paths for the \acp{UAV} to cooperatively gather sensor data subject to constraints on energy, time, or distance. 
Our motivation is to allow the \acp{UAV} to \textit{adaptively} monitor the terrain in areas of interest where information value is high.

Traditional approaches for data collection are \textit{non-adaptive} and rely on static, predefined paths. 
In multi-\ac{UAV} coverage path planning~\citep{Bahnemann2017}, the terrain is equally partitioned and a sweep pattern is assigned to each \ac{UAV}. 
The main drawback of such methods is that they assume uniformly distributed target variables of interest, e.g. anomalies, hotspots, victims, and do not allow for targeted inspection of specific areas. To address this, \ac{IPP} methods~\citep{Hollinger2013,Albani2018,Tzes2022,Ruckin2021,Viseras2019,bayerlein2021multi} have been proposed enabling online decision-making based on incoming information. 
However, the runtime of these strategies usually scales exponentially with the planning horizon, since they rely on evaluating many candidate paths online.
In team scenarios, reasoning about the other \acp{UAV}' behaviours causes the number of evaluations to further grow exponentially with the team size, which leads to intractable runtime complexity.

 \begin{figure}[!t]
    \centering
    \includegraphics[width=\columnwidth]{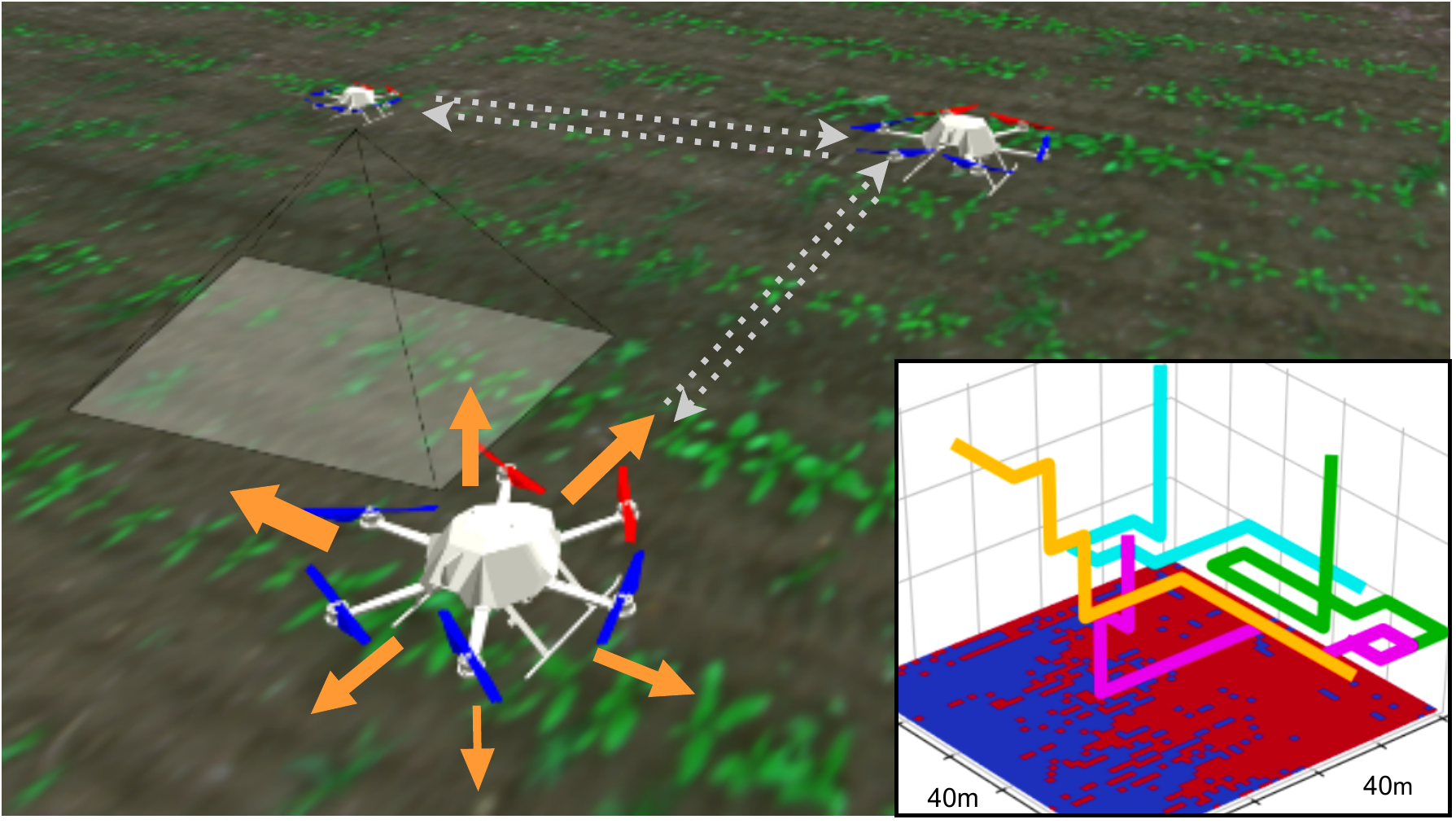}
    \caption{Our RL-based approach applied in a multi-\ac{UAV} surface temperature mapping scenario. The \acp{UAV} take images of the terrain (white transparent) and communicate them (grey dashed arrows). Based on locally available information, each \ac{UAV} decides from a set of actions (orange arrows) where to take the next measurement. Inset image: resulting trajectories for $4$ deployed \acp{UAV} (different colours). By planning paths cooperatively, our approach enables adaptively mapping warm (red) areas of interest on the field.} \label{F:teaser}
    \vspace{-3mm}
\end{figure}

\Ac{RL} has emerged as a popular approach for efficiently learning online decision-making in robotics~\citep{Ruckin2021,Chen2020,Fan2020,Viseras2019,bayerlein2021multi,Luis2022}.
Recent works apply \ac{RL} for single-agent \ac{IPP} to enhance path quality and computation time for adaptive data collection~\citep{Chen2020,Ruckin2021,Cao2022,Lodel2022}.
However, learning informative paths for multiple cooperating agents remains an open challenge.
First studies show promising results~\citep{Viseras2019,bayerlein2021multi}, but are limited to 2D action spaces and do not address the credit assignment problem~\citep{Agogino2004}, i.e. how much each agent contributes to the overall team performance, which adversely impacts cooperation capabilities.

The main contribution of this paper is a novel multi-agent deep \ac{RL}-based \ac{IPP} approach for adaptive terrain monitoring scenarios using \ac{UAV} teams. 
Bridging the gap between recent advances in \ac{RL} and robotic applications, we build upon \ac{COMA}~\citep{Foerster2018} to explicitly address the credit assignment problem in cooperative \ac{IPP}. 
Our approach supports decentralised on-board decision-making and achieves cooperative 3D path planning with variable team sizes.
We show that (i) our designed network input representations are effective for multi-agent IPP in 3D action spaces;
(ii) our COMA-based algorithm accounts for credit assignment resulting in improved planning performance; and
(iii) our RL-based approach improves planning performance compared to non-learning-based methods across varying team sizes and communication constraints without re-training.
To back up these claims, we demonstrate the performance of our approach using synthetic and real-world data in a thermal hotspot mapping scenario.
We will open-source our code at:
\texttt{https://github.com/dmar-bonn/ipp-marl}.

\section{Related Work} \label{S:related_work}

Our work brings together multi-robot \ac{IPP} for autonomous data collection and advances in multi-agent \ac{RL}. This section overviews relevant literature for both, distinguishing between single- and multi-robot settings.

\textbf{Informative path planning} has been widely studied for efficient data gathering using autonomous robots~\citep{Hollinger2013,Albani2018,Viseras2019,Popovic2020,bayerlein2021multi,Carbone2021,Ruckin2021,Cao2022}.
\textit{Non-adaptive} coverage planners~\citep{Bahnemann2017} monitor a terrain exhaustively based on pre-defined paths. In contrast, we focus on \textit{adaptive} strategies replanning paths online based on incoming sensor data.
\citet{Hollinger2013} study the benefit of adaptive online replanning to maximise the information value of sensor measurements given a limited mission budget for single-vehicle underwater inspection tasks.
Recent works~\citep{Popovic2020,Blanchard2022} propose optimisation-based \ac{IPP} methods for single-\ac{UAV} monitoring.
However, these methods require computationally costly evaluations of candidate paths' expected information value. Extending them naïvely to multiple \acp{UAV} yields exponentially scaling complexity which restricts applicability to online applications.

Deploying cooperative robotic teams is beneficial for monitoring tasks over large terrains such as post-disaster assessment~\citep{Goodrich2008,Hayat2017} and precision agriculture~\citep{Albani2018,Carbone2021,Maes2019}. 
We focus on decentralised planning strategies, where robots make locally informed decisions on-board to enable robustness to individual failures and scalability with the team size.
\citet{Best2019} introduce a decentralised variant of Monte Carlo tree search over a joint probability distribution of action sequences to plan individual paths for active perception.
Similar to our multi-\ac{UAV} setup, \citet{Albani2018} and \citet{Carbone2021} propose methods for monitoring areas of interest in crop fields.
The former use biology-inspired swarm behavior for adaptive \ac{IPP}. Despite promising results, their method relies on manual problem-specific parameter tuning.

An alternative line of work leverages learning-based methods for active data collection.
Following the paradigm of centralised training and decentralised policy execution, \citet{Li2020} learn inter-robot communication policies to exchange local robot information. \citet{Tzes2022} propose a more general multi-robot framework relying on graph neural networks.
Both approaches are trained via supervised imitation learning and require an expert to learn from.
We learn the desired team behaviour using deep \ac{RL}, which enables flexible, foresighted planning in various applications.

\textbf{Reinforcement learning (RL)} is increasingly utilised in robotics and \ac{UAV} applications~\citep{Chen2020,Ruckin2021, Cao2022, Lodel2022,Pirinen2022}. \citet{Pirinen2022} introduce a strategy for finding an unknown goal region using a \ac{UAV} based on limited visual cues. Recently, \ac{RL} has also been applied to \ac{IPP} to efficiently replan robotic paths online. \citet{Chen2020} develop a graph-based deep \ac{RL} method for exploration, selecting map frontiers that reduce map uncertainty and travel time. 
However, their approach is limited to 2D workspaces, while we consider 3D planning. Other works reward agent actions that lead to high information gain~\citep{Lodel2022} and map uncertainty reduction in target regions~\citep{Cao2022}. 
In a similar problem setup to ours, \citet{Ruckin2021} propose an \ac{IPP} method combining deep \ac{RL} and sampling-based planning for adaptive \ac{UAV} terrain monitoring in 3D workspaces.
Naïvely extending single-agent algorithms to multi-agent setups incurs exponentially growing complexity with the number of agents.

Multi-agent \ac{RL} for \ac{IPP} is a relatively unexplored research area.
Existing approaches for cooperative team applications do not account for the credit assignment problem~\citep{Agogino2004}, i.e. do not discriminate the contribution of one agent to the overall team performance.
Most works address related applications including navigation~\citep{Fan2020}, target assignment~\citep{Qie2019}, and coverage planning~\citep{Puente2022}. 
Similar to us, \citet{bayerlein2021multi} propose a multi-agent \ac{RL}-based \ac{IPP} approach that maximises harvested data without inter-agent communication. 
\citet{Viseras2019} allow agents to exchange information via a communication module.
Both works are limited to constant \ac{UAV} altitudes ignoring potentially varying sensor noise with altitude. 
Recently, \citet{Luis2022} proposed a deep Q-learning algorithm that supports learning cooperation by penalising redundant measurements based on the inter-\ac{UAV} distance. Their reward design is tailored to pure exploration instead of adaptively monitoring areas of interest.

These works independently assign hand-engineered individual agent rewards.
Thus, they do not fully account for the cooperative nature of \ac{IPP} and require manual reward tuning.
In contrast, we propose a new approach solely relying on generally applicable global rewards for the whole \ac{UAV} team.
We adapt the counterfactual multi-agent (COMA) \ac{RL} algorithm~\citep{Foerster2018} to active robotic data collection in 3D workspaces. This way, we explicitly assign credit to individual agents during training. Our experimental results emphasise the need for explicit credit assignment to achieve cooperative behaviour and verify that it improves \ac{IPP} performance.

\section{Problem Statement} \label{S:problem_statement}

We consider a team of homogeneous \acp{UAV} monitoring a flat terrain.
The goal is to plan information-rich \ac{UAV} paths on-the-fly as new measurements arrive to accurately and efficiently map regions of interest on the terrain given a finite mission budget.
We briefly describe the general multi-agent \ac{IPP} problem, our mapping strategy, and how to quantify information value for our \ac{RL} approach.

\subsection{Multi-Agent Informative Path Planning}
\label{sec:IPP}

We address the problem of multi-agent \acf{IPP} optimising an information-theoretic criterion $\mathrm{I}: \Psi^{N} \to \mathbb{R}^{+}$ over all $N$ \ac{UAV} paths $\psi = \{\psi_{1}, \ldots, \psi_{N}\}$, where $\Psi$ is the set of all possible individual \ac{UAV} paths:
\begin{equation}\label{eq:IPP}
    \psi^{*} = \argmax_{\psi \in \Psi^{N}} \mathrm{I}(\psi),\, \text{s.t. } \mathrm{C}(\psi_i) \leq B\ \forall\ i \in \{1, \ldots,N\}\,. 
\end{equation}

Each set of \ac{UAV} paths $\psi$ is composed of individual paths $\psi_{i} = (\bm{p}_i^0, \ldots, \bm{p}_i^B) \in \mathrm{\Psi}$ of length $B$ with 3D measurement positions $\bm{p}_i^t \in \mathbb{R}^3$ in an equi-distant grid $\mathrm{P}$ of resolution $r_P$ over multiple altitudes above the terrain. The function $\mathrm{C}: \Psi \to \mathbb{R}^{+}$ maps a \ac{UAV} path $\psi_i$ to its associated execution cost; in our work, a maximum number $B \in \mathbb{N}^{+}$ of measurements taken along a path $\psi_i$.

\begin{figure}[!t]
    \centering
    \includegraphics[width=0.45\textwidth]{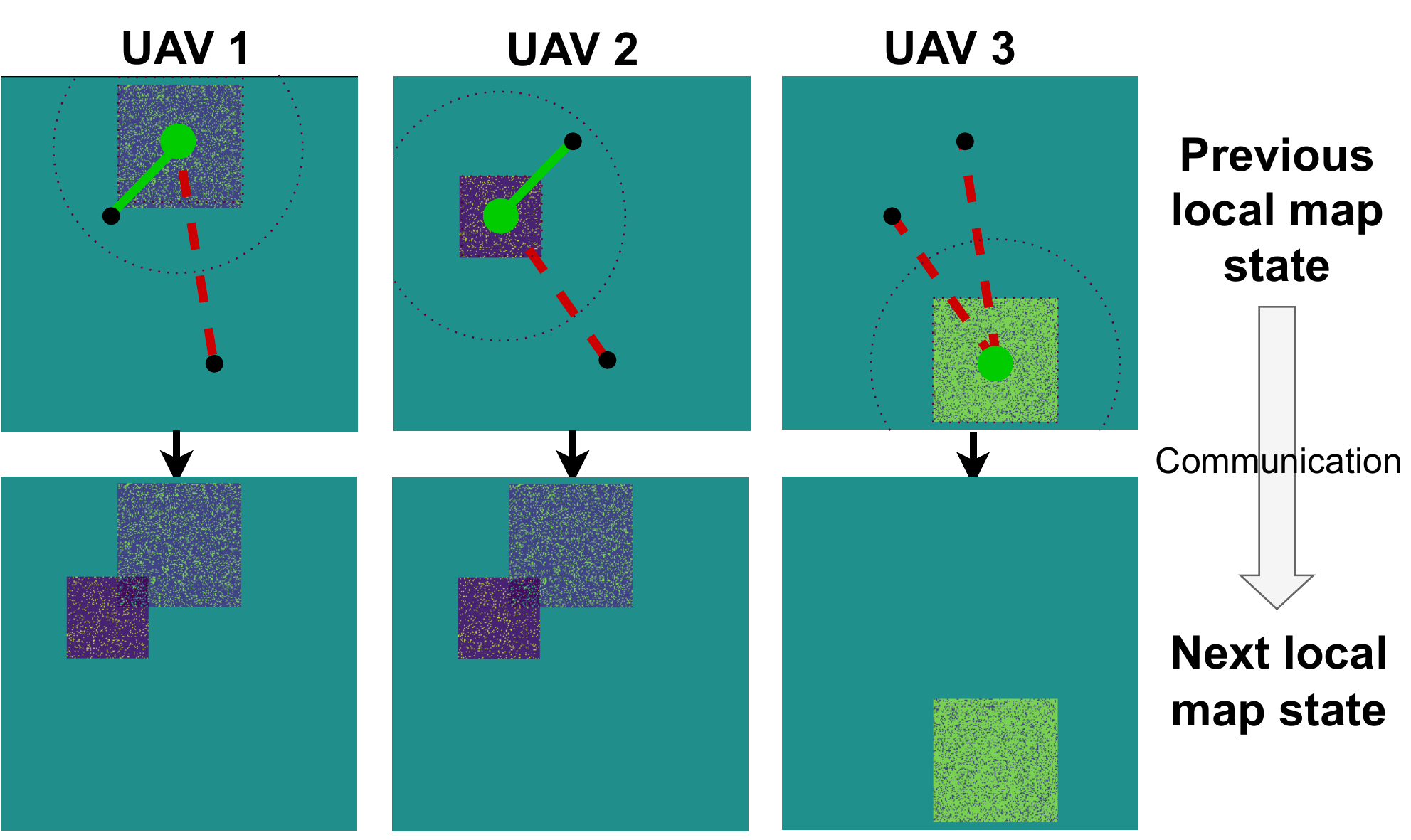}
    \caption{Inter-\ac{UAV} communication. The \acp{UAV} (filled circles) exchange their current measurements (square footprints) with each other when closer than a limited communication range (black dotted circle). Green and dotted red lines indicate in-range and out-of-range communications, respectively. The \acp{UAV} use the received measurements to update their local map states (top to bottom row).}
    \label{F:communication}
    \vspace{-3mm}
\end{figure}

\subsection{Terrain Mapping}

The \acp{UAV} map the terrain by taking images using downward-facing cameras. The image information is processed, e.g. by semantic segmentation, communicated, and fused into a terrain map.
A measurement $z_i^t$ taken by \ac{UAV} $i$ at time step $t$ is a likelihood estimate projected onto the flat terrain. 
We consider binary per-pixel classification and sequentially fuse $z_i^t$ using probabilistic occupancy grid mapping~\citep{Elfes1989}.
Each \ac{UAV} $i$ stores a local posterior map belief $\mathcal{\bm{M}}_{i}$ discretised into $M$ grid cells $\mathcal{M}_{i}^{j}$ of resolution $r_{M}$. 
The altitude of the current measurement position $\bm{p}_i^t$ determines the mapped field of view and the mapping resolution.
We align $r_{M}$ with the mapping resolution at the lowest altitude and upsample higher-altitude measurements to $r_{M}$ to fuse heterogeneous mapping resolutions together in $\mathcal{M}_{i}$.
Similar to~\citet{Popovic2020}, we leverage an altitude-dependent sensor model specifying $p(z_i^t \,|\, \mathcal{M}_{i}^j, \bm{p}_{i}^t)$ to update a \ac{UAV}'s local posterior map belief over each $\mathcal{M}_{i}^{j}$ at time step $t$.

\Cref{F:communication} visualises the inter-\ac{UAV} communication protocol. All \acp{UAV} $i$ send their measurements $z_i^t$ to another \ac{UAV} $k$ and receive measurements $z_k^t$ to update the individual local map beliefs over $\mathcal{M}_i$.
\acp{UAV} $i$ and $k$ share measurements, if $\lVert \bm{p}_i^t - \bm{p}_k^t \rVert_2 \leq D$, where $D \in \mathbb{R}^{+}$ is a radius approximating a range-limited communication channel.
The communication message contains the \ac{UAV} identifier $i$, its position $\bm{p}_i^t$, and its collected measurement $z_i^t$. 
The receiving \ac{UAV} $k$ utilises $z_i^t$ and $p_i^t$ to update its local map belief over $\mathcal{M}^{k}$.

\subsection{Utility Definition for Adaptive Terrain Monitoring}

Our goal is to plan future measurement positions $\psi_i^{t+1} = (\bm{p}_i^{t+1}, \ldots, \bm{p}_i^{B})$ for each \ac{UAV} $i$ at time step $t$ from which the next measurements $\{z_i^{t+1}, \ldots, z_i^B\}$ maximally reduce the uncertainty of the current posterior belief over a global map $\mathcal{M}$. The global map $\mathcal{M}$ contains measurements $z_i^{0:t}$ taken by the $N$ \acp{UAV} up to the current time step $t$. We quantify the uncertainty reduction of a set of measurements $z^{t+1} = \{z_1^{t+1}, \ldots, z_N^{t+1}\}$ taken at the next time step $t+1$ by computing the entropy reduction over $\mathcal{M}$ after fusing $z^{t+1}$. Then, the information criterion $\mathrm{I}(\psi^{t+1})$ is computed as the summed entropy reduction along paths $\psi^{t+1}$:
\begin{equation}\label{eq:entropy_reduction}
    \begin{aligned}
        \mathrm{I}(\psi^{t+1}) = \sum_{m=0}^{B-t-1} &H(\mathcal{M} \,|\, z^{0:t+m}, \bm{p}^{0:t+m}) - \\
        &H(\mathcal{M} \,|\, z^{0:t+m+1}, \bm{p}^{0:t+m+1})\,.
    \end{aligned}
\end{equation}

We consider adaptive mapping with one interesting target class, i.e. one class that holds information value, and a complement class.
Thus, the map entropy $H(\mathcal{M} \,|\, z^{0:t}, \bm{p}^{0:t})$ at time step $t$ is weighted by the importances of the interesting and the uninteresting classes:
\begin{equation}\label{eq:entropy}
    \begin{aligned}
        &H(\mathcal{M} \,|\, z^{0:t}, \bm{p}^{0:t}) = \sum_{j=1}^M H(\mathcal{M}^j \,|\, z_{0:t}, \bm{p}_{0:t}) \\
        = & -\sum_{j=1}^{M} \mathrm{W}(\mathcal{M}^j) \, p(\mathcal{M}^j \,|\, z^{0:t}, \bm{p}^{0:t}) \, \log\big(p(\mathcal{M}^j \,|\, z^{0:t}, \bm{p}^{0:t}\big) \\
        &+ \mathrm{W}(\overline{\mathcal{M}^j}) \, p(\overline{\mathcal{M}^j} \,|\, z^{0:t}, \bm{p}^{0:t}) \, \log\big(p(\overline{\mathcal{M}^j} \,|\, z^{0:t}, \bm{p}^{0:t})\big)\,,
    \end{aligned}
\end{equation}
where $p(\overline{\mathcal{M}^j} \,|\, z^{0:t}, \bm{p}^{0:t}) = 1 - p(\mathcal{M}^j \,|\, z^{0:t}, \bm{p}^{0:t})$. The importance weighting $W(\mathcal{M}^j)$ is defined as:
\begin{equation}\label{eq:weighting_function}
    W(\mathcal{M}^j) = 
    \begin{cases}
        w_1 & \text{if } p(\mathcal{M}^j \,|\, z^{0:t}, \bm{p}^{0:t}) > 0.5\,, \\
        w_2 & \text{if } p(\mathcal{M}^j \,|\, z^{0:t}, \bm{p}^{0:t}) < 0.5\,, \\
        0.5 & \text{else} \, ,
    \end{cases}
\end{equation}
where $w_1, w_2 \geq 0$ and $w_1 + w_2 = 1$. The importance weights $w_1$, $w_2$ encourage the \acp{UAV} to plan paths $\psi^{t+1}$ targeting potentially interesting regions according to the current posterior map belief $p(\mathcal{M}^j \,|\, z^{0:t}, \bm{p}^{0:t})$.

\section{Our Approach} \label{S:approach}

\begin{figure}[!t]
    \centering
    \includegraphics[width=0.45\textwidth]{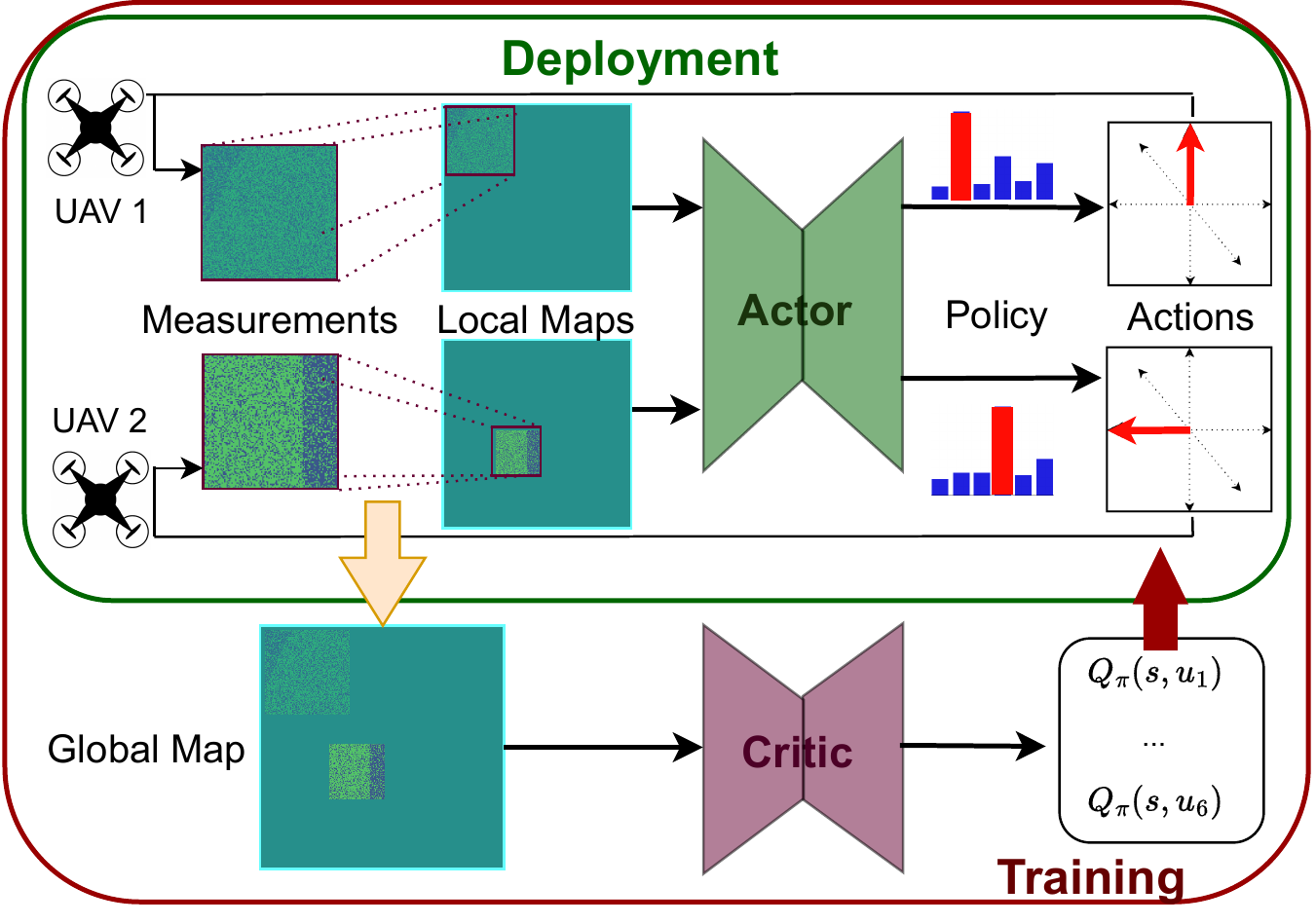}
    \caption{Overview of our approach. At each time step during a mission, each \ac{UAV} takes a measurement and updates its local map state. The local map is input to an actor network, which outputs a policy from which an action is sampled. During training, a centralised critic network is additionally trained using global map information and outputs Q-values for each action from the current state, i.e. the expected future return.}
    \label{F:algorithm_overview}
    \vspace{-3mm}
\end{figure}

We present our novel multi-agent \ac{RL}-based \ac{IPP} approach for \ac{UAV} teams. 
Our goal is to plan \ac{UAV} paths in a 3D workspace to achieve cooperative adaptive terrain mapping. As shown in \Cref{F:algorithm_overview}, we train agents offline based on global terrain information using \ac{RL} to learn \ac{UAV} paths in a centralised way. 
A key aspect is the integration of a counterfactual baseline, allowing us to estimate each agent's mapping contribution to the overall team performance and improve cooperation.
During a mission, we leverage the trained agent behaviour and deploy a fully decentralised system to replan informative measurement positions online.

\subsection{RL for Sequential Decision-Making}
\label{sec:rl}

We formulate the multi-\ac{UAV} \ac{IPP} task as a sequential decision-making problem for a team of agents and train it using \ac{RL}.
The \acp{UAV} execute missions, where at each time step $t$ each \ac{UAV} $i$ simultaneously plans its next measurement position $\bm{p}_{i}^{t+1}$ based on the current local on-board state $\omega_i^t$. 
The local state $\omega_i^t$ includes the local map belief $\mathcal{M}_i$ that sequentially fuses past (communicated) measurements $z^{0:t}$ encoding the measurement history. This way, we account for partial observability induced by communication constraints and noisy sensor measurements.
At each time step, all \ac{UAV} actions, i.e. next measurement positions, define the joint action $\bm{u}^t = (u_1^t, \ldots, u_N^t) \in U^N$. The \acp{UAV} receive one global team reward $R^t: S \times U^N \times S \rightarrow \mathbb{R}$ quantifying the joint information value of mapped measurements $z^{t+1} = \{z_1^t, \ldots, z_N^t\}$. The discounted return $G^t = \sum_{k=0}^{B-t} \gamma^k R^{t+k}$, where $\gamma \in [0,1)$ is a discount factor, measures the information value of the team's paths from $t$ until the mission budget is spent. Although each \ac{UAV} chooses actions in a decentralised way, we evaluate the performance of the team as a whole to enforce cooperative \ac{IPP} behaviour.

\noindent \textbf{State}. The global environment state $s^t \in S$ is defined as $s^t = \{\mathcal{M}^t, \bm{p}_{1:N}^{t}, b\}$. $ \mathcal{M}^t$ captures the current global map belief $p(\mathcal{M} \,|\, z^{0:t}_{1:N}, \bm{p}^{0:t}_{1:N})$, $\bm{p}_{1:N}^{t}$ are the current agent positions, and $b \leq B$ is the remaining mission budget. $s^{t+1}$ is the next state after the agents have moved to the next positions $\bm{p}_{1:N}^{t+1}$ reached by executing actions $\bm{u}^t$. $ \mathcal{M}^{t+1}$ is updated based on measurements $z^{t+1}$ taken at $\bm{p}_{1:N}^{t+1}$.

\noindent \textbf{Actions}. The agents move within a discrete position grid $\mathrm{P}$, bounded by the environment borders and discrete minimum and maximum altitudes. 
Each \ac{UAV} $i$ selects an action $u_i$ from a discrete 3D action space $U$ containing movements $\{up, north, east, south, west, down\}$ with a fixed step size.
We prevent actions leading to movement outside of the environment or \acp{UAV} having concurrent 2D terrain coordinates.

\noindent \textbf{Reward}.  
We explicitly design the reward function $R$ to reflect the information criterion defined in \Cref{eq:entropy_reduction} to adaptively and quickly reduce map uncertainty in target areas:
\begin{equation}\label{eq:reward}
R^t(s^t, \bm{u}^t, s^{t+1}) = \alpha \, \frac{H(\mathcal{M}^{t+1}) - H(\mathcal{M}^{t})\,}{H(\mathcal{M}^{t})} + \beta\,.
\end{equation}

We reward the weighted map entropy reduction from the current to the next map state $\mathcal{M}^{t+1}$ after mapping new measurements $z^{t+1}$. We normalise the reward by the current map entropy to keep its magnitude approximately constant during the mission and apply affine scaling factors $\alpha$ and $\beta$ to improve training stability. 
Note that, for $\gamma = 1$, the return $G^t$ resembles the \ac{IPP} criterion in \Cref{eq:entropy_reduction} up to the normalisation and scaling factors. 

\subsection{Algorithm}
\label{S:algorithm}

Our goal is to learn a policy enabling cooperative \ac{UAV} team behaviour for the adaptive monitoring task.
To do this, we build upon the \ac{COMA} \ac{RL} algorithm of~\citet{Foerster2018} to our \ac{IPP} application for \ac{UAV} teams. 
\ac{COMA} is an actor-critic algorithm using a centralised critic network to evaluate each agent's behaviour, described by the policy $\pi(\cdot \,|\, \omega_i^t)$ and parameterised by an actor network, and to optimise the policy accordingly.
The critic evaluates the current policy $\pi$ by estimating the agent's $i$ state-action value $Q_{\pi}\big(s^t,(u_1^t, \ldots, u_i^t, \ldots, u_N^t)\big)$ given the other agents' actions $\bm{u}_{-i}^t$.
The critic network is trained on-policy via TD($\lambda$)~\citep{Sutton2018} to estimate the discounted return $G_t$, introduced in \Cref{sec:rl}, for taking the joint action $\bm{u}^t$ from the current state $s^t$ and following the current policy $\pi$ afterwards.
The critic uses global information $s^t$ during training, while the actor utilises only on-board information $\omega_i^t$ to predict the next-best measurement position decentralised during both training and deployment.
We leverage the counterfactual baseline to assign credit to individual agents according to their contribution to the team performance, which fosters cooperative team behaviour. 
The advantage $A_i^t$ for agent $i$ taking action $u_{i}^{t}$ in the team's action $\bm{u}^t$ is:
\begin{equation}\label{eq:COMA}
\small{A_{i}^t(s^t, \bm{u}^t) = Q_{\pi}(s^t, \bm{u}^t) - \sum_{{u'}_{i}^t \in U} \pi({u'}_{i}^t \,|\, \omega_{i}^t) \, Q_{\pi}\big(s^t, (\bm{u}_{-i}^t, {u'}_{i}^t)\big)\,.}
\end{equation}

The contribution of agent $i$ to the joint state-action value $Q_{\pi}(s^t,\bm{u}^t)$ of the team's action $\bm{u}^t = (u_1^t, \ldots, u_i^t, \ldots, u_N^t)$ by taking action $u_{i}^t$ is estimated by marginalising over all possible individual actions ${u'}_{i}^t \in U$ while keeping the other agents' actions $\bm{u}_{-i}^t$ fixed. 
For optimising $ \pi(\cdot \,|\, \omega_{i}^t)$, we apply the policy gradient theorem using \Cref{eq:COMA} and minimise:
\begin{equation}\label{eq:actor_loss}
\mathcal{L} = -\log \pi(u_i^t|\omega_{i}^t) \, A_i^t(s^t, \bm{u}^t)\,,
\end{equation}
using mini-batch stochastic gradient descent.
For further details, we refer to \citet{Foerster2018}.

\subsection{Network Architecture \& Feature Design}
\label{S:networks}

\begin{figure}[!t]
    \centering
    \includegraphics[width=0.42\textwidth]{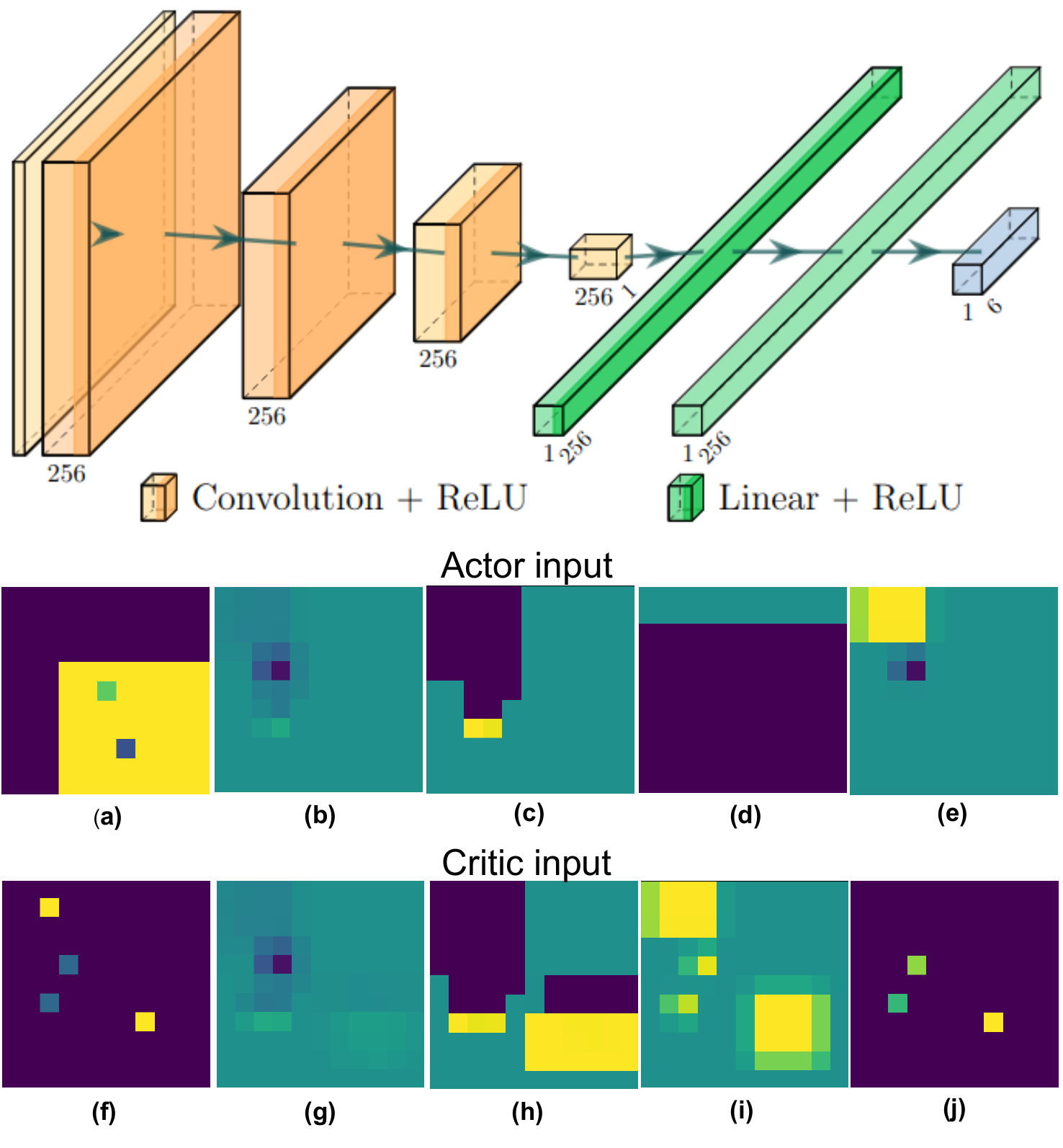}
    \caption{Architecture and inputs for our actor and critic. Both networks consist of convolutional encoders and linear layers in the prediction head. For the actor, the output is produced by a softmax layer (grey).
    Both actor and critic receive local inputs (`Actor input'). The critic also receives global inputs for centralised training (`Critic input'). All inputs (a)-(j) are detailed in \Cref{S:networks}.}   
    \label{F:network}
    \vspace{-3mm}
\end{figure}

We propose new actor and critic representations to exploit \ac{COMA} in 3D robotic applications.
As shown in \Cref{F:network}, actor and critic are represented by neural networks $f_{\theta^{\pi}}$ and $f_{\theta^{c}}$, parameterised by $\theta^{\pi}$ and $\theta^{c}$.
The actor network is conditioned on the agent $i$ and its local state $\omega_i^{t}$.
Feature inputs are the agent's identifier $i$, the remaining mission budget $b$, and the following spatial inputs: (a) a position map centred around the agent's position encoding the map boundaries and the communicated other agents' positions, where the values represent the agent altitudes; (b) the local map state $\mathcal{M}_i$; (c) the weighted entropy of the local map state $H(\mathcal{M}_i \,|\, z^{0:t}, \bm{p}^{0:t})$ in \Cref{eq:entropy}; (d) the weighted entropy of the measurement $H(z_{i}^{t} \,|\, \bm{p}_{i}^{t})$; and (e) the map cells currently spanned by all agents' fields of view within the communication range ('footprint map').

Our critic network receives the same input (a)-(e) and, in addition, global environment information accessible during training. Specifically, it further receives: (f) a global position map encoding all agent positions $\bm{p}_{1:N}^t$; (g) the global map state $\mathcal{M}$; (h) its weighted entropy $H(\mathcal{M} \,|\, z^{0:t}, \bm{p}^{0:t})$; (i) the map cells currently spanned by all agents' fields of view; and, to enable learning the counterfactual baseline in \Cref{eq:COMA}, (j) the other agents' actions.
Inputs are provided in the position grid resolution $r_P$. We downsample the map resolution $r_M$ by $\frac{r_P}{r_M}$ to align both resolutions.
The scalar inputs $i$ and $b$ are expanded to constant-valued 2D feature maps. 

To handle spatial information necessary for the terrain monitoring task, the networks consist of convolutional encoders and multi-layer perceptron heads predicting the policy $\pi$ and Q-values $Q_{\pi}$, respectively.
\Cref{F:network}-\textit{Top} illustrates the architecture of both networks.
The actor's logits $f_{\theta^{\pi}}(x^t_i)$ given a collection $x^t_i$ of feature maps (a)-(e) above, are passed through a bounded softmax function $\pi(u_{i} \,|\, x^t_i) = (1 - \epsilon) \, \mathrm{softmax}\big(f_{\theta^{\pi}}(x^t_i)\big) + \frac{\epsilon}{|U|}$ to predict the stochastic policy.
The hyperparameter $\epsilon \in [0,1]$ fosters exploration during training and is set to $0$ at deployment. The critic outputs Q-values $Q_{\pi}\big(s^t, (u_1^t, \ldots, u_i^t, \ldots, u_N^t)\big)$ for each action $u_i^t \in U$ of agent $i$ 
after the last linear layer.

\subsection{Network Training}
\label{S:Training}

We train an actor and a critic network offline and utilise the trained actor online at deployment.
We simulate \ac{UAV} missions and learn from rewards received after fusing measurements $z$ into the global map $\mathcal{M}$.
Before each mission, we generate a new terrain of resolution $r_M$. 
The terrains are split into connected interesting and complementary uninteresting regions. 
The split is randomly oriented and interesting regions cover between $30\%$ to $60\%$ of the terrain to foster generalisation.
We fix the initial \ac{UAV} positions and execute a mission until budget $B$ is spent.
During training, actions are sampled from the actor's policy as described in \Cref{S:networks}, where $\epsilon$ is linearly decreased from $0.5$ to $0.02$ over the first $10,000$ missions.

We alternate between generating $3,000$ environment interactions on-policy and policy optimisation using \Cref{eq:actor_loss}. 
Both networks are optimised via stochastic mini-batch gradient descent for $5$ epochs using the Adam optimiser with learning rates of $1e^{-5}$ (actor) and $1e^{-4}$ (critic), and a batch size of $600$.
The critic network is trained to estimate the expected return $G_t$ applying TD($\lambda$) with $\lambda=0.8$ and $\gamma=0.99$ using a target critic network copying the critic network each $30,000$ environment interactions.

\section{Experimental Results} \label{S:results}

We present experiments to show the capabilities of our multi-agent \ac{RL}-based \ac{IPP} approach for adaptive terrain monitoring using \ac{UAV} teams.
Our experimental results support our claims, which are: (i) our designed network representations are effective for multi-agent \ac{IPP} for \acp{UAV} in a 3D workspace; (ii) accounting for the credit assignment problem via a counterfactual baseline improves planning performance; and (iii) our approach outperforms non-learning-based state-of-the-art approaches in terms of planning performance across varying team sizes and communication constraints. Moreover, we demonstrate our approach applied to a real-world surface temperature monitoring scenario.

\subsection{Experimental Setup} \label{s:experimental_setup}

For each experiment, we execute $50$ terrain monitoring missions with changing regions of interest as described in \Cref{S:Training}. 
The terrains are of size $50$\,m $\times$ $50$\,m with a map resolution of $r_M = 10$\,cm.
We set the planning resolution to $r_P = 5$\,m, bound altitudes between $5$\,m and $15$\,m, and use camera field of views of $60^{\circ}$, so that adjacent measurements do not overlap when taken from the lowest altitude.
To account for increased sensor noise at higher altitudes, we simulate $p(z_i^t \,|\, \mathcal{M}_{i}^j, \bm{p}_{i}^t)$ to be $\{0.99, 0.735, 0.625\}$ at $\{5, 10, 15\}$\,m altitude.
The \ac{UAV} team consists of $4$ agents and the communication radius is limited to $25$\,m unless reported otherwise.
As metrics, we use the entropy of the map state $H(\mathcal{M} \,|\, z^{0:t}, \bm{p}^{0:t})$ to assess the map uncertainty and the F1-score between the map state $\mathcal{M}$ and the ground truth map to evaluate the correctness of $\mathcal{M}$.
We report the mean and standard deviation of these performance metrics in ground truth regions of interest given $B=15$ measurements. 

\begin{figure}[!t]
    \centering
    \includegraphics[height=0.265\textwidth]{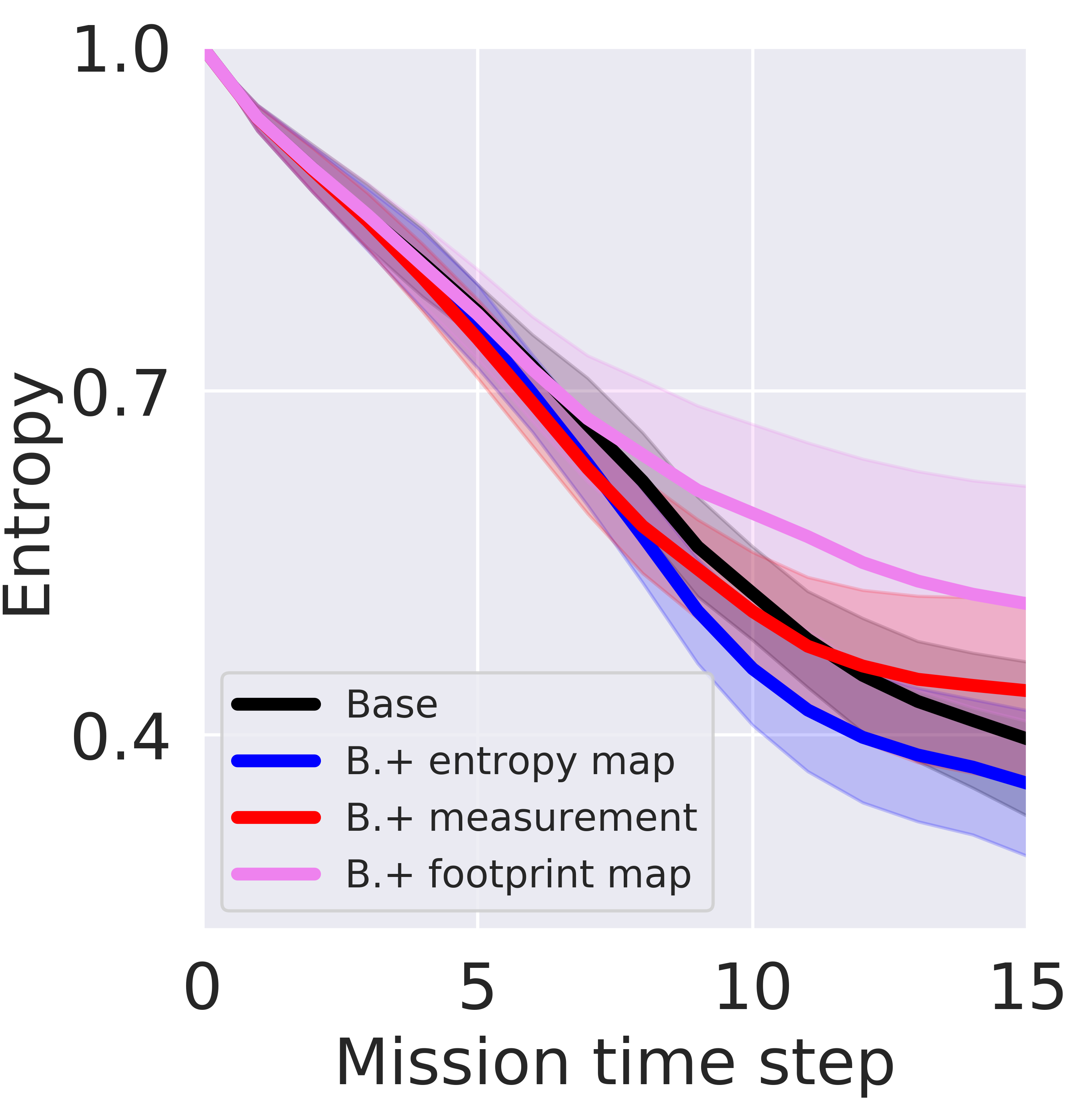}
    \includegraphics[height=0.265\textwidth]{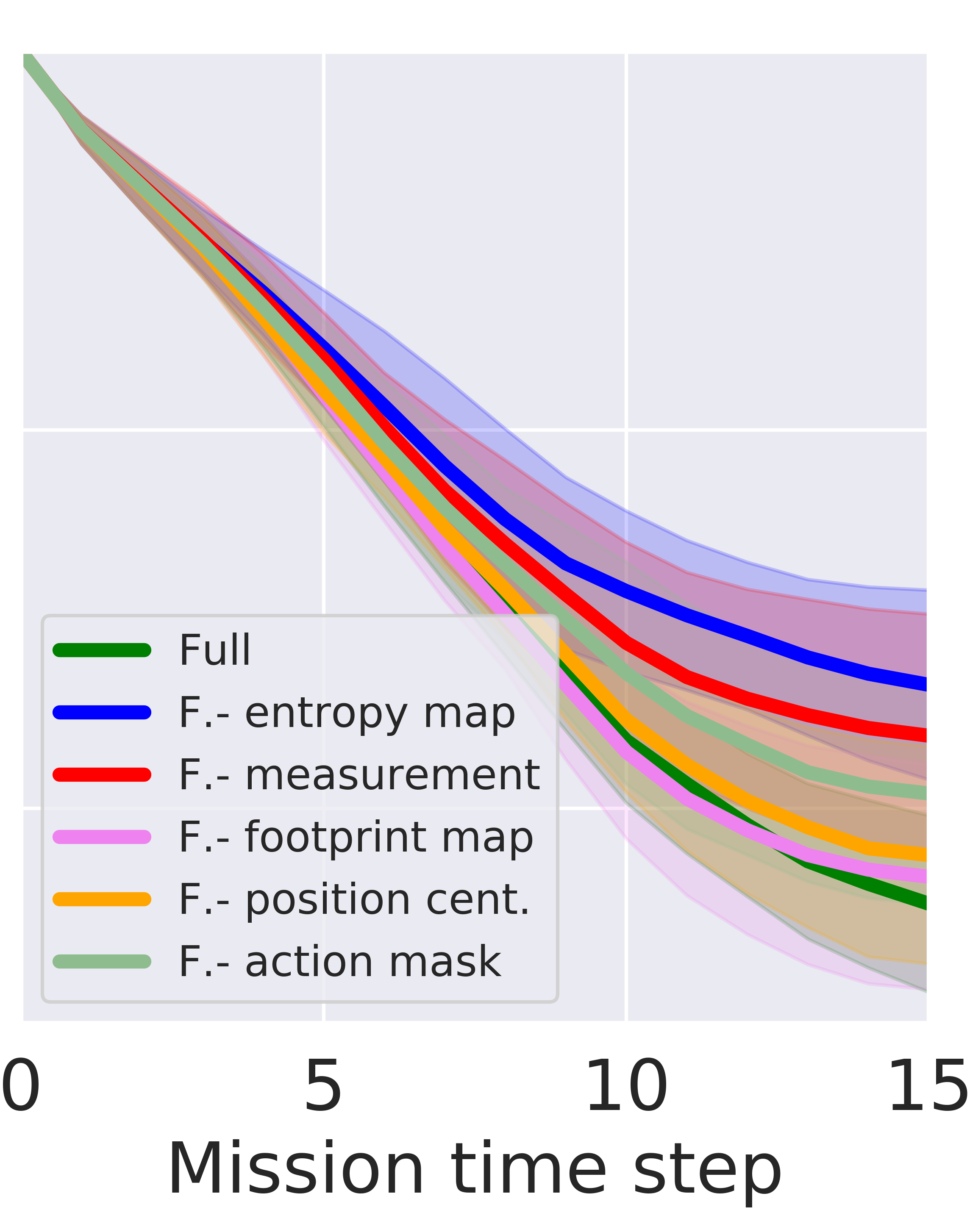}
    \caption{Feature input ablations. We systematically add (left) and remove (right) components of our approach and show the map entropy reduction over the mission time. The results show that the entropy map input has the largest impact, while the footprint map only contributes to a better performance at the end of a mission.
    Our full proposed network input (green) performs best.}
    \label{F:feature_ablation}
    \vspace{-3mm}
\end{figure}

\subsection{Ablation Study of Network Feature Design} \label{s:feature_ablation}

In this section, we provide ablations to show that our proposed network input representation is effective for multi-agent \ac{IPP} for \acp{UAV} in a 3D workspace. 
Our ablation studies systematically add and remove single input feature maps to assess their effect on the overall planning performance. 

\Cref{F:feature_ablation} shows the mean map entropy reduction and standard deviation during the mission with varying network input features. Steeper falling curves indicate better planning performance.
In \Cref{F:feature_ablation}\textit{-Left}, we add single input feature maps to \textit{base} input features (a)-(d), as described in \Cref{S:networks}, which are necessary to model the \ac{IPP} problem.
In \Cref{F:feature_ablation}\textit{-Right}, we remove single input feature maps.

As expected, the proposed full input (green) performs best, i.e. leads to the lowest final map entropy.
The entropy map feature (blue) is most beneficial for planning performance, 
suggesting that it holds most relevant information for learning informative paths. Interestingly, adding the footprint map (pink) slightly harms planning performance, but leads to worse final performance when removed from the input. 
This indicates that, when combined with the other inputs, the footprint map has a larger contribution at the end of the mission.
Intuitively, this is because identifying the currently observed map cells, which is essential for planning the next measurement positions, becomes harder as more measurements are mapped.
Adding the current measurement's entropy (red) decreases uncertainty faster during the first $\sim 12$ mission time steps but harms final performance, presumably since this local feature causes myopic planning bias.
As part of the full input with enough global features, however, this feature is significant for planning performance providing additional map information in close proximity.
Additionally, not masking invalid actions (light green) and not centring the agents' position map (orange) impairs performance since this removes valuable spatial information for planning.
In sum, the results confirm that our network input features (`Full') are most effective for multi-agent \ac{IPP} for \acp{UAV} in a 3D workspace, leading to superior planning performance of our proposed \ac{RL}-based approach compared to possible variants.

\subsection{Credit Assignment Mechanism Study} \label{s:coma_ablation}

The experiments in this section show that accounting for the credit assignment problem results in improved planning performance, which verifies the need for explicit credit assignment mechanisms in cooperative \ac{RL}-based multi-agent \ac{IPP}. 
We perform a systematic study comparing the \ac{UAV} team's \ac{IPP} performance with varying advantage functions in \Cref{eq:COMA} and thus changing policy gradient updates in \Cref{eq:actor_loss}. 
To this end, we compare our proposed approach (\Cref{S:algorithm}) against three variants of itself, as described in the following.

\begin{figure}[!t]
    \centering
    \includegraphics[width=0.235\textwidth]{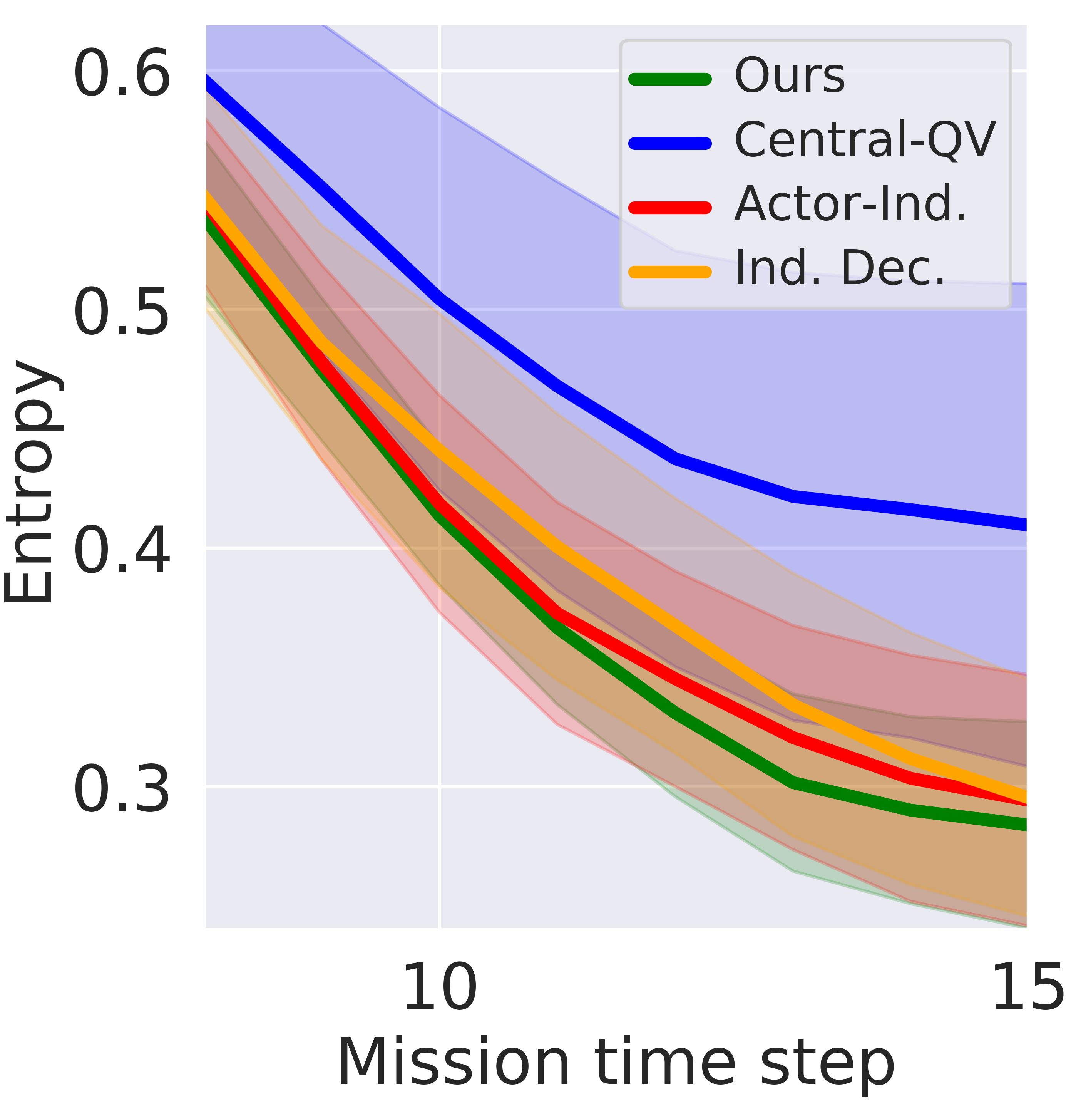}
    \includegraphics[width=0.235\textwidth]{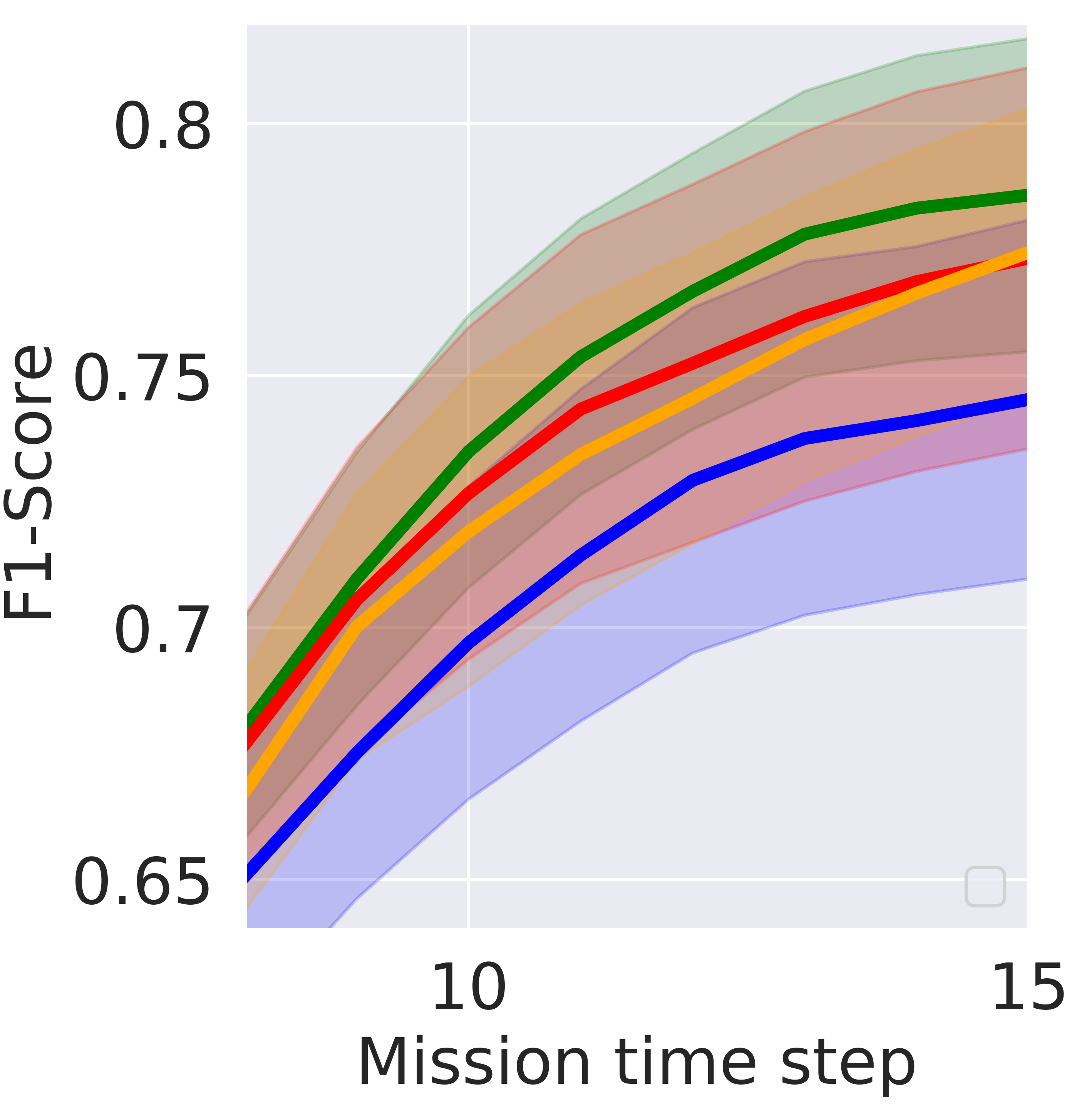}
    \caption{Credit assignment study. We compare our COMA-approach (green) with variants which do not explicitly tackle the credit assignment problem in terms of map entropy (left) and F1-score (right). In later mission stages, our approach allows for better cooperation for adaptive mapping when most of the environment has already been explored.}
    \label{F:coma_ablation}
    \vspace{-3mm}
\end{figure}

\noindent \textbf{Central-QV.} 
Similar to \citet{Foerster2018}, we verify the effectiveness of the counterfactual baseline in our adaptive \ac{IPP} scenario by replacing the baseline in \Cref{eq:COMA} with a state value $V_{\pi}(s^t)$. This way, we do not estimate an individual agent's contribution but consider the team's joint performance alone. 
We train two critic networks estimating $Q$ and $V$, both as described in \Cref{S:Training}, and change the advantage function to $A_{i}^{t}(s^t, \bm{u}^t) = Q_{\pi}(s^t, \bm{u}^t) - V_{\pi}(s^t)$.

\noindent \textbf{Actor-Independent.} As described in \Cref{S:algorithm}, we utilise a centralised critic exploiting global state information during training. However, to investigate the effect of reasoning about the other team members' actions, we now do not account for them and exclude them from the critic network input. We adapt the advantage function to depend solely on the agent's own action: $A_{i}^{t}(s^t, u_{i}^{t}) = Q_{\pi}(s^t, u_{i}^{t}) - \sum_{{u}_{i}^{'t} \in U} \pi (u_{i}^{'t} \,|\, \omega_{i}^{t}) Q_{\pi}(s^t, u_{i}^{'t})$.

\noindent \textbf{Decentralised.} In this variant, we remove all global information and consider a purely decentralised critic based on local agent information $\omega_i^t$ only. As for the \textit{actor-independent} variant, the critic ignores the other agents' actions using the own agent's state value as a baseline.
The advantage function reduces to: $A_{i}^{t}(\omega_{i}^{t}, u_{i}^{t}) = Q_{\pi}(\omega_{i}^{t}, u_{i}^{t}) - \sum_{{u}_{i}^{'t} \in U} \pi (u_{i}^{'t} \,|\, \omega_{i}^{t}) Q_{\pi}(\omega_{i}^{t}, u_{i}^{'t})$.

\Cref{F:coma_ablation} shows the planning performance using different advantage functions. We focus on performance at later stages of the mission, when most of the environment is explored and cooperation is crucial for adaptive mapping.
The counterfactual baseline utilised in our approach (green) performs best, indicating its effectiveness for achieving cooperative behaviour.
All variants lack explicit credit assignment mechanisms, which adversely impacts planning performance irrespective of using centralised or decentralised critics.
This confirms that our \ac{COMA}-based algorithm improves planning performance by addressing the credit assignment problem.

\begin{figure}[!t]
    \centering
    \includegraphics[width=0.235\textwidth]{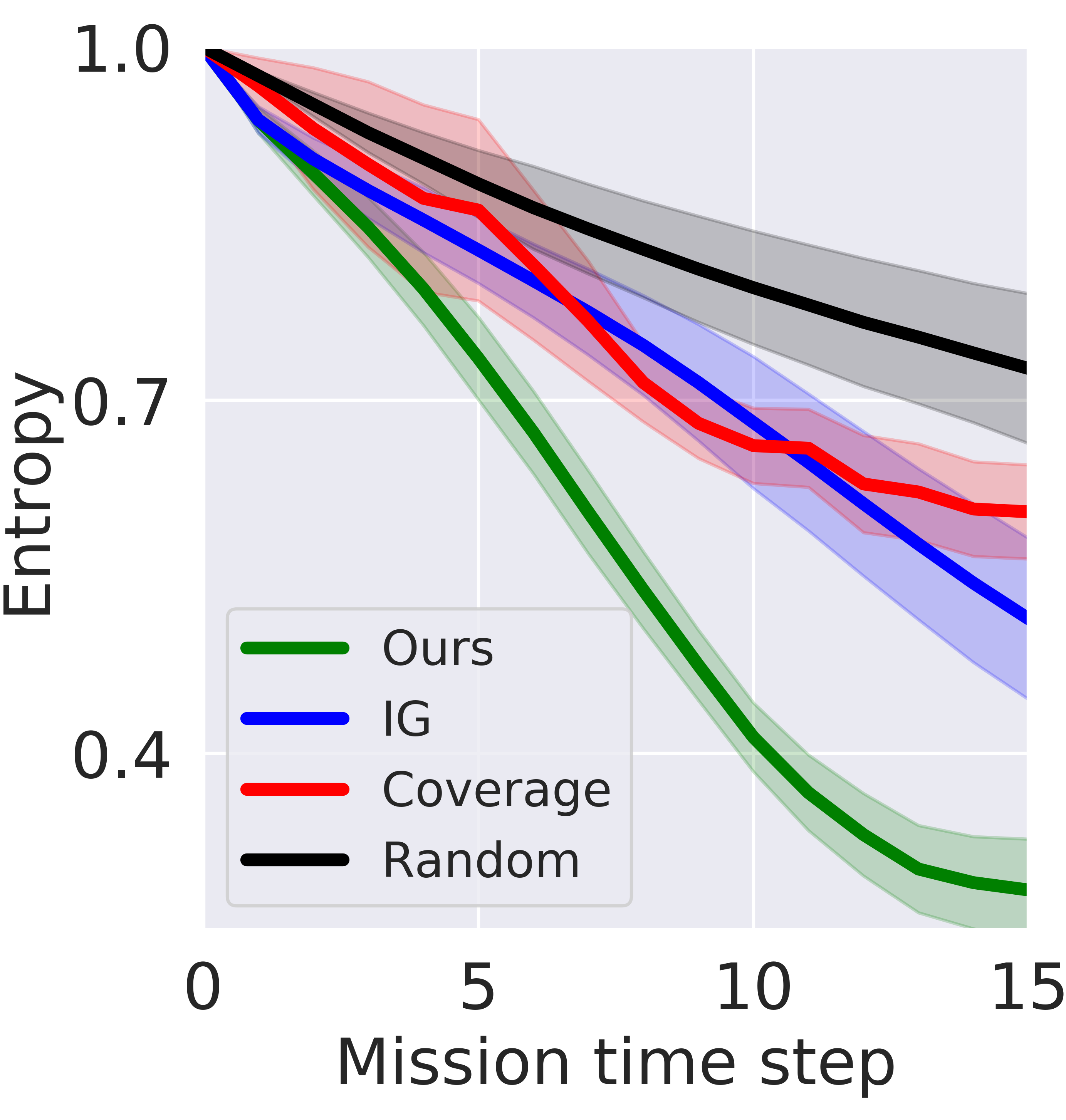}
    \includegraphics[width=0.235\textwidth]{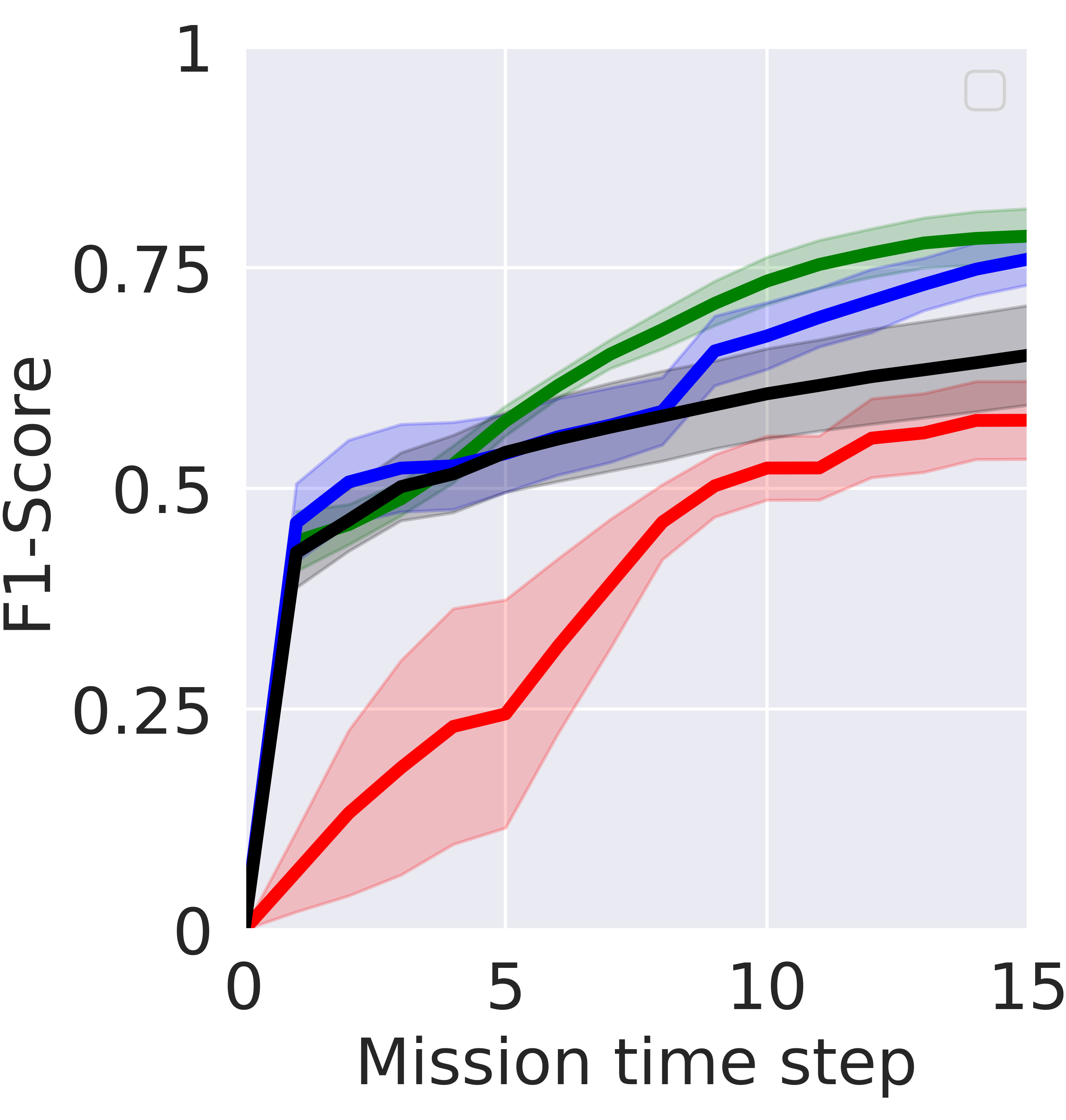}
    \caption{Comparison of planning methods for $4$ agents and a communication range of $25m$. By planning paths cooperatively, our approach reduces map uncertainty (left) and improves map accuracy (right) quickest, indicating its superior performance.}
    \label{F:baseline_comparison}
    \vspace{-3mm}
\end{figure}

\begin{table*}[h]
\caption{Robustness to varying team sizes and communication constraints. We report the mean and standard deviation of entropy and F1-score over $50$ trials after $33\%$, $66\%$, and $100\%$ of the mission time. Best results in bold.} \label{T:robustness}
\centering
\setlength{\tabcolsep}{3.1pt}
\begin{tabular}{@{}clcccccc@{}}
Setting      &   Approach                     & 33\% Entropy $\downarrow$ & 67\% Entropy $\downarrow$ & 100\% Entropy $\downarrow$ & 33\% F1 $\uparrow$ & 67\% F1 $\uparrow$ & 100\% F1 $\uparrow$ \\ \midrule
2 agents  &  Ours      & $\textbf{0.8826} \pm 0.0267$                & $\textbf{0.7343} \pm 0.0548$                   & $\textbf{0.5802} \pm 0.0451$                   & $\textbf{0.3608} \pm 0.0661$                   & $\textbf{0.5035} \pm 0.0768$                  & $\textbf{0.6268} \pm 0.0436$                   \\
   &   IG            & $0.9155 \pm 0.0154$                   & $0.8449 \pm 0.0410$                  & $0.7432 \pm 0.0801$                   & $0.3275 \pm 0.0386$                 & $0.4353 \pm 0.0539$                & $0.5340 \pm 0.0857$               \\ 
    &  Coverage       & $0.9163 \pm 0.0472$                   & $0.7845 \pm 0.0628$                  & $0.6970 \pm 0.0332$                   & $0.1603 \pm 0.0871$                 & $0.3687 \pm 0.0912$                & $0.4863 \pm 0.0418$               \\ \hline \noalign{\vskip 1mm}
4 agents   &   Ours    & $\textbf{0.7360} \pm 0.0349$                  & $\textbf{0.4137} \pm 0.0294$                   & $\textbf{0.2842} \pm 0.0408$                   & $\textbf{0.5765} \pm 0.0168$                  & $\textbf{0.7350} \pm 0.0268$                  &  $\textbf{0.7858} \pm 0.0308$                  \\
   &   IG    & $0.8302 \pm 0.0298$                   & $0.6805 \pm 0.0554$                & $0.5176 \pm 0.0700$                   & $0.5396 \pm 0.0435$                  &  $0.6728 \pm 0.0375$                  &  $0.7599 \pm 0.0289$                 \\  
    &   Coverage     & $0.8615 \pm 0.0762$                   & $0.6614 \pm 0.0317$                & $0.6052 \pm 0.0396$                   & $0.1603 \pm 0.0870$                  &  $0.3687 \pm 0.0913$                  &  $0.4864 \pm 0.0418$                 \\ \hline \noalign{\vskip 1mm} 
8 agents   &  Ours        & $\textbf{0.5621} \pm 0.0235$                   & $\textbf{0.2952} \pm 0.0403$                   & $\textbf{0.2209} \pm 0.0459$                   & $\textbf{0.7115} \pm 0.0336$                  & $0.7859 \pm 0.0322$                  & $0.8171 \pm 0.0283$                \\
    &  IG           & $0.6887 \pm 0.0234$                   & $0.4886 \pm 0.0460$                   & $0.3077 \pm 0.0446$                   & $0.6957 \pm 0.0260$                  & $\textbf{0.7881} \pm 0.0194$                 & $\textbf{0.8576} \pm 0.0151$                  \\  
    &  Coverage       & $0.8185 \pm 0.0832$                   & $0.6072 \pm 0.0274$                   & $0.5149 \pm 0.0309$                   & $0.2455 \pm 0.1281$                  & $0.5246 \pm 0.0358$                 & $0.5800 \pm 0.0433$                  \\  \noalign{\vskip 1mm} \hline \hline \noalign{\vskip 1mm}

Zero communication  & Ours       & $\textbf{0.7638} \pm 0.0408$                & $\textbf{0.5445} \pm 0.0674$                   & $\textbf{0.3699} \pm 0.0486$                   & $\textbf{0.5476} \pm 0.0256$                   & $\textbf{0.6846} \pm 0.0327$                  & $\textbf{0.7620} \pm 0.0282$                   \\
  &  IG        & $0.8284 \pm 0.0283$                   & $0.6814 \pm 0.0529$                  & $0.5286 \pm 0.0775$                   & $0.5396 \pm 0.0453$                 & $0.6510 \pm 0.0442$                & $0.7151 \pm 0.0432$               \\ \hline \noalign{\vskip 1mm}
Limited communication  &  Ours     & $\textbf{0.7360} \pm 0.0349$                  & $\textbf{0.4137} \pm 0.0294$                   & $\textbf{0.2842} \pm 0.0408$                   & $\textbf{0.5765} \pm 0.0168$                  & $\textbf{0.7350} \pm 0.0268$                  &  $\textbf{0.7858} \pm 0.0308$                  \\
   &   IG    & $0.8302 \pm 0.0298$                   & $0.6805 \pm 0.0554$                & $0.5176 \pm 0.0700$                   & $0.5396 \pm 0.0435$                  &  $0.6728 \pm 0.0375$                  &  $0.7599 \pm 0.0289$                 \\  \hline \noalign{\vskip 1mm}
Full communication   &  Ours       & $\textbf{0.7428} \pm 0.0276$                   & $\textbf{0.4623} \pm 0.0602$                   & $\textbf{0.3674} \pm 0.0843$                   & $\textbf{0.5818} \pm 0.0294$                  & $\textbf{0.7077} \pm 0.0481$                  & $0.7524 \pm 0.0465$                \\
   &  IG        & $0.8294 \pm 0.0290$                   & $0.6809 \pm 0.0501$                   & $0.5266 \pm 0.0593$                   & $0.5389 \pm 0.0449$                  & $0.6721 \pm 0.0432$                 & $\textbf{0.7606} \pm 0.0306$                  \\ 
\end{tabular}
\vspace{-3mm}
\end{table*}

\subsection{Comparison against Non-Learning-based Approaches} \label{s:baseline_comparison}

The following experiments back up our claim that our \ac{RL}-based approach outperforms state-of-the-art non-learning-based multi-agent \ac{IPP} approaches. 
We compare our approach against three methods:
(i) an adaptive information gain approach for \ac{UAV} swarms proposed by~\citet{Carbone2021}, which selects a \ac{UAV}'s action greedily by maximising the estimated map entropy reduction without explicit credit assignment (`IG'); 
(ii) a non-adaptive coverage lawnmower-like pattern with equidistant ($5$\,m) measurement positions at the best-performing altitude (`Coverage');
(iii) non-adaptive random exploration sampling \ac{UAV} actions uniformly at random (`Random'). 
All approaches share the same state space, action space, and action masking strategy.

\Cref{F:baseline_comparison} shows the evaluation metrics over the mission time for the considered approaches.
The information gain-based strategy (blue) and our \ac{RL}-based approach (green) outperform the non-adaptive methods as they can actively focus on regions of interest.
The superior performance of our approach confirms the benefits of addressing the credit assignment problem and the applicability of our learning-based planning to varying terrains.

Next, we show the generalisability of our learning-based approach to different team sizes and communication settings. We deploy our actor trained on the $4$-agent setting with communication limited to $25$\,m in scenarios with $2$, $4$, and $8$ agents, and communication radii of $0$\,m, $25$\,m, and unlimited communication without re-training.
\Cref{T:robustness} shows the planning performance of our approach compared to the information gain-based and the coverage method.
As expected, more agents lead to faster mapping, i.e. entropy decrease and F1-score increase.
Though changing communication radii leads to small performance drops, our approach consistently outperforms both.
This verifies that our \ac{RL} approach generalises to varying numbers of agents and communication requirements without re-training, showcasing its broad applicability without additional training costs.

\subsection{Temperature Mapping Scenario} \label{s:real_world_results}

\begin{figure}[!t]
    \centering
    \begin{minipage}[c]{0.235\textwidth}
    \includegraphics[width=\textwidth]{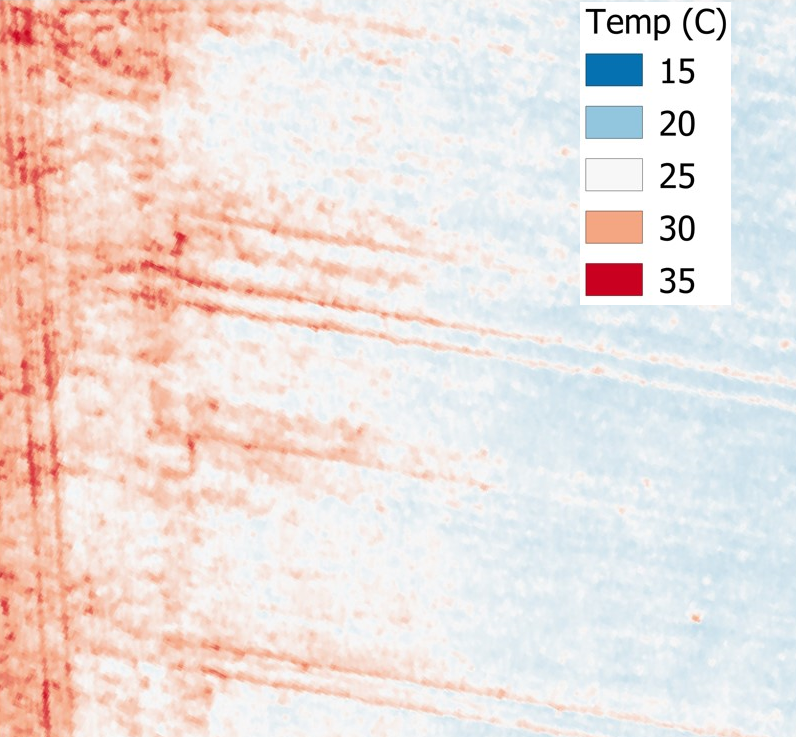}
    \end{minipage}
    \begin{minipage}[c]{0.235\textwidth}
    \includegraphics[width=\textwidth]{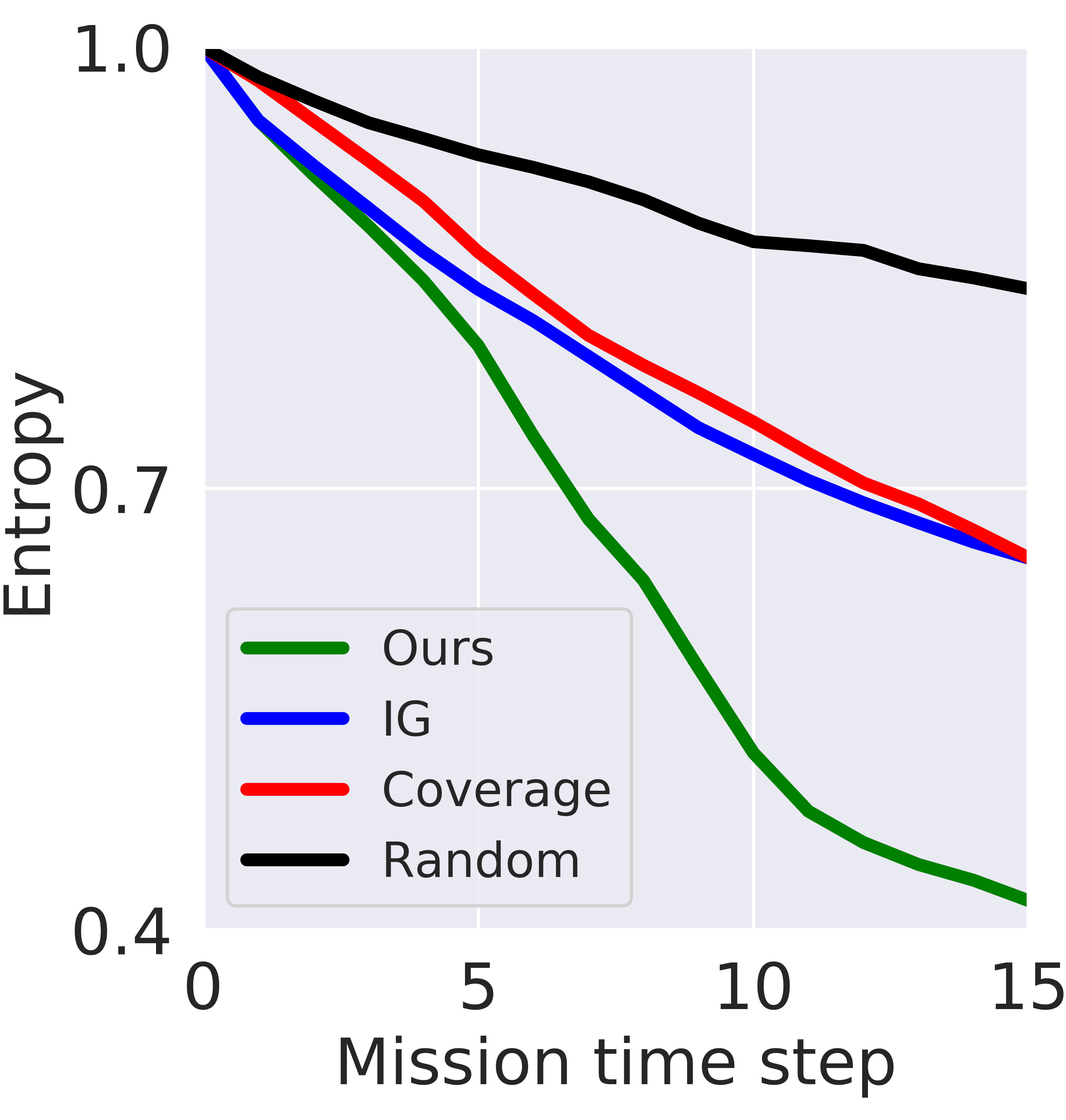}
    \end{minipage}
    \caption{Real-world evaluation. We deploy our approach on surface temperature data (left) and compare the map entropy reduction of the considered planning methods. Warmer regions are considered as interesting. Although our approach is trained on synthetic data, it outperforms all other methods. \Cref{F:teaser} shows the resulting \ac{UAV} paths planned in this experiment.}
    \label{F:real_world}
    \vspace{-3mm}
\end{figure}

We demonstrate the performance of our approach in a surface temperature monitoring scenario using real-world data of a $40$\,m $\times 40$\,m crop field near J\"{u}lich, Germany. The field data was collected with a DJI Matrice 600 \ac{UAV} carrying a Vue Pro R 640 thermal sensor and is shown in \Cref{F:real_world}-\textit{Left}. 
We discretise the terrain into a $500 \times 500$ grid map with resolution $r_M = 8$\,cm. 
The planning grid resolution is $r_P = 4$\,m to guarantee the same network input dimensions as used for training.
Regions with surface temperature $\geq 25^{\circ}$\,C are considered as being interesting for adaptive hotspot mapping.
\Cref{F:real_world}-\textit{Right} shows the map entropy reduction over $B=15$ measurements for $4$ agents and a communication radius of $25$\,m, using our approach compared to the methods introduced in \Cref{s:baseline_comparison}. 
Note that our approach is trained solely in simulation on synthetic data as described in \Cref{S:Training}. Our approach clearly outperforms all other methods showcasing its applicability for real-world terrain monitoring missions.

\section{Conclusions and Future Work} \label{S:conclusions}

In this paper, we introduced a novel multi-agent deep \ac{RL}-based \ac{IPP} approach for adaptive terrain monitoring using \ac{UAV} teams. 
Our method features new network representations and exploits a counterfactual baseline to address the credit assignment problem for cooperative \ac{UAV} path planning in a 3D workspace.
This allows us to successfully outperform state-of-the-art non-learning-based approaches in terms of monitoring efficiency, while generalising to different mission settings without re-training.
Experiments using \ac{UAV}-acquired thermal data validate the real-world applicability of our approach.
Future work will investigate heterogeneous robot teams and dynamically growing maps in environments of unknown bounds.

\bibliographystyle{IEEEtranN}
\footnotesize
\bibliography{2023-westheider-icra}

\end{document}